\documentclass[11pt,a4paper]{article}

\usepackage[hyperref]{emnlp-ijcnlp-2019}
\usepackage{times}
\usepackage{latexsym}
\usepackage{url}
\usepackage{todonotes}
\usepackage{enumitem}
\usepackage{array}
\usepackage[justification=centering]{caption}
\usepackage{graphicx}
\usepackage{subcaption}
\usepackage{multicol}

\makeatletter
\newcommand{\printfnsymbol}[1]{%
  \textsuperscript{\@fnsymbol{#1}}%
}
\makeatother

\aclfinalcopy 

\newcommand\BibTeX{B{\sc ib}\TeX}
\newcommand\confname{EMNLP-IJCNLP 2019}
\newcommand\conforg{SIGDAT}

\title{Incorporating Domain Knowledge into Medical NLI\\ using Knowledge Graphs}
\author{Soumya Sharma\thanks{equal contribution} \\ \texttt{soumyasharma20\textsuperscript{$\dagger$}} \And
Bishal Santra\printfnsymbol{1} \\ \texttt{bsantraigi\textsuperscript{$\dagger$}} \And
Abhik Jana \\ \texttt{abhikjana1\textsuperscript{$\dagger$}}
\AND 
T Y S S Santosh \\ \texttt{santoshtyss\textsuperscript{$\dagger$}} \And
Niloy Ganguly \\ \texttt{niloy\textsuperscript{$\ddagger$}} \And
Pawan Goyal \\ \texttt{pawang\textsuperscript{$\ddagger$}} 
\AND
\\
Indian Institute of Technology Kharagpur, India \\
\texttt{\{$\dagger$\}@gmail.com, \{$\ddagger$\}@cse.iitkgp.ac.in}}

\date{}
\begin{document}
\maketitle
\begin{abstract}
Recently, biomedical version of embeddings obtained from language models such as BioELMo have shown state-of-the-art results for the textual inference task in the medical domain. In this paper, we explore how to incorporate structured domain knowledge, available in the form of a knowledge graph (UMLS), for the Medical NLI task. Specifically, we experiment with fusing embeddings obtained from knowledge graph with the state-of-the-art approaches for NLI task, which mainly rely on contextual word embeddings. We also experiment with fusing the domain-specific sentiment information for the task. Experiments conducted on MedNLI dataset clearly show that this strategy improves the baseline BioELMo architecture for the Medical NLI task\footnote{\url{https://github.com/soummyaah/KGMedNLI/}}.  
\end{abstract}

\section{Introduction}
Natural language inference (NLI) is one of the basic natural language understanding tasks which deals with detecting inferential relationship such as entailment or contradiction, between a given premise and a hypothesis. In recent years, with the availability of large annotated datasets like SNLI~\cite{bowman-etal-2015-large}, MultiNLI~\cite{williams2018broad}, researchers have come up with several neural network based models which could be trained with these large annotated datasets and are able to produce state-of-the-art performances~\cite{bowman-etal-2015-large,bowman2016fast, munkhdalai2017neural,sha2016reading,chen2017enhanced,tay2017compare}. With these attempts, even though NLI in domains like fiction, travel etc. has progressed a lot,
NLI in medical domain is yet to be explored extensively. With the introduction of MedNLI~\cite{romanov-shivade-2018-lessons}, an expert annotated dataset for NLI in the clinical domain, researchers have started pursuing the problem of clinical NLI. Modeling informal inference is one of the basic tasks towards achieving natural language understanding, and is considered very challenging. 
MedNLI is a dataset that assists in assessing how good a sentence or word embedding method is for downstream uses in medical domain.

Recently, with the emergence of advanced contextual word embedding methods like ELMo~\cite{peters2018deep} and BERT~\cite{devlin2018bert}, performances of many NLP tasks have improved, setting state-of-the-art performances. 
Following this stream of literature, ~\citet{lee2019biobert} introduce BioBERT, which is a BERT model pre-trained on English Wikipedia, BooksCorpus and fine-tuned on PubMed (7.8B tokens in total) corpus, PMC full-text articles. ~\citet{jin2019probing} propose BioELMo which is a domain-specific version of ELMo trained on 10M PubMed abstracts, and attempt to solve medical NLI problem with these domain specific embeddings, leading to state-of-the-art performance. These two attempts show a direction towards solving medical NLI problem where the pretrained embeddings are fine-tuned on medical corpus and are used in the state-of-the-art NLI architecture.

~\citet{chen2018neural} proposed the use of external knowledge to help enrich neural-network based NLI models by applying Knowledge-enriched co-attention, Local inference collection with External Knowledge, and Knowledge-enchanced inference composition components. Another line of solution tries to bring in the extra domain knowledge from sources like Unified Medical Language System (UMLS)~\cite{bodenreider2004unified}. ~\citet{romanov-shivade-2018-lessons} used the knowledge-directed attention based methods in ~\cite{chen2018neural} for Medical NLI. Another such attempt is made by~\citet{lu2019incorporating}, where they incorporate domain knowledge in terms of the definitions of medical concepts from UMLS with the state-of-the-art NLI model ESIM~\cite{chen2017enhanced} and vanilla word embeddings of Glove~\cite{pennington2014glove} and fastText~\cite{bojanowski2017enriching}. Even though, the authors achieve significant improvement by incorporating only concept definitions from UMLS, the features of this clinical knowledge graph are yet to be fully utilized. Motivated by the emerging trend of embedding knowledge graphs to encode useful information in a high dimensional vector space, we propose the idea of applying state-of-the-art knowledge graph embedding algorithm on UMLS and use these embeddings as a representative of additional domain knowledge with the state-of-the-art medical NLI models like BioELMo, to investigate the performance improvement on this task. Additionally, we also incorporate the sentiment information for medical concepts given by MetaMap ~\cite{metamap} leading to further improvement of the performance.  
Note that, as state-of-the-art baselines, we use the models proposed by~\citet{jin2019probing} and~\citet{lu2019incorporating}. Since, both of these studies consider ESIM as the core NLI model which makes it more convenient for us to incorporate extra domain knowledge and to have a fair performance comparison with these state-of-the-art models.        
Our contributions are two-fold.
\begin{itemize}
    \item We incorporate domain knowledge via knowledge graph embeddings applied on UMLS. We propose an intelligent path-way to combine contextual word embeddings with the domain specific embeddings learnt from the knowledge graph, and feed them to the state-of-the-art NLI models like ESIM. This helps to improve the performance of the base architecture.
    \item We further show the usefulness of the associated sentiments per medical concept from UMLS in boosting the performance further, which in a way shows that if we can carefully use the domain knowledge present in sources like UMLS, it can lead to promising results as far as the medical NLI task is concerned. 
\end{itemize}

\section{Dataset}
In this study, we use the MedNLI dataset ~\cite{romanov-shivade-2018-lessons}, a well-accepted dataset for natural language inference in clinical domain. The dataset is sampled from doctors' notes in the clinical dataset MIMIC-III~\cite{mimic-iii} and is arguably the largest publicly available database of patient records. The entire dataset consists of 14,049 premise-hypothesis pairs divided into 11,232 train pairs, 1,395 validation pairs and 1,422 test pairs. Each such pair consists of a gold label which could be either entailment (true), contradiction (false), or neutral (undetermined). The average (maximum) sentence lengths of premises and hypotheses are 20 (202) and 5.8 (20), respectively.

\section{Proposed Approach}
The task is to classify the given premise ($p$) and hypothesis ($h$) sentence pair into one of the three classes: entailment, contradiction and neutral. Following the approach by~\citet{jin2019probing}, we use the BioELMo embedding model where authors bring in contextual information in terms of embeddings obtained via applying ELMo trained on 10M PubMed abstracts, and use these with the state-of-the-art ESIM model~\cite{chen2017enhanced} for the NLI task. The architecture includes two sentence encoders each of which takes in as input the word embeddings of $p$ and $h$. The inputs are run through corresponding bi-directional LSTM encoder layers. Pairwise attention matrix is computed between encoded $p$ and $h$, which forms the attention layer followed by a second bi-directional LSTM layer run separately over $p$ and $h$. Max and average pooling are performed over the outputs of LSTM layers and the output of pooling operations is run through a softmax model. 
We feed this architecture an additional domain knowledge from UMLS as vector representations obtained via knowledge graph embeddings, the details of which are described below. \\   

\begin{figure}[t]
  \centering
    \includegraphics[width=\linewidth]{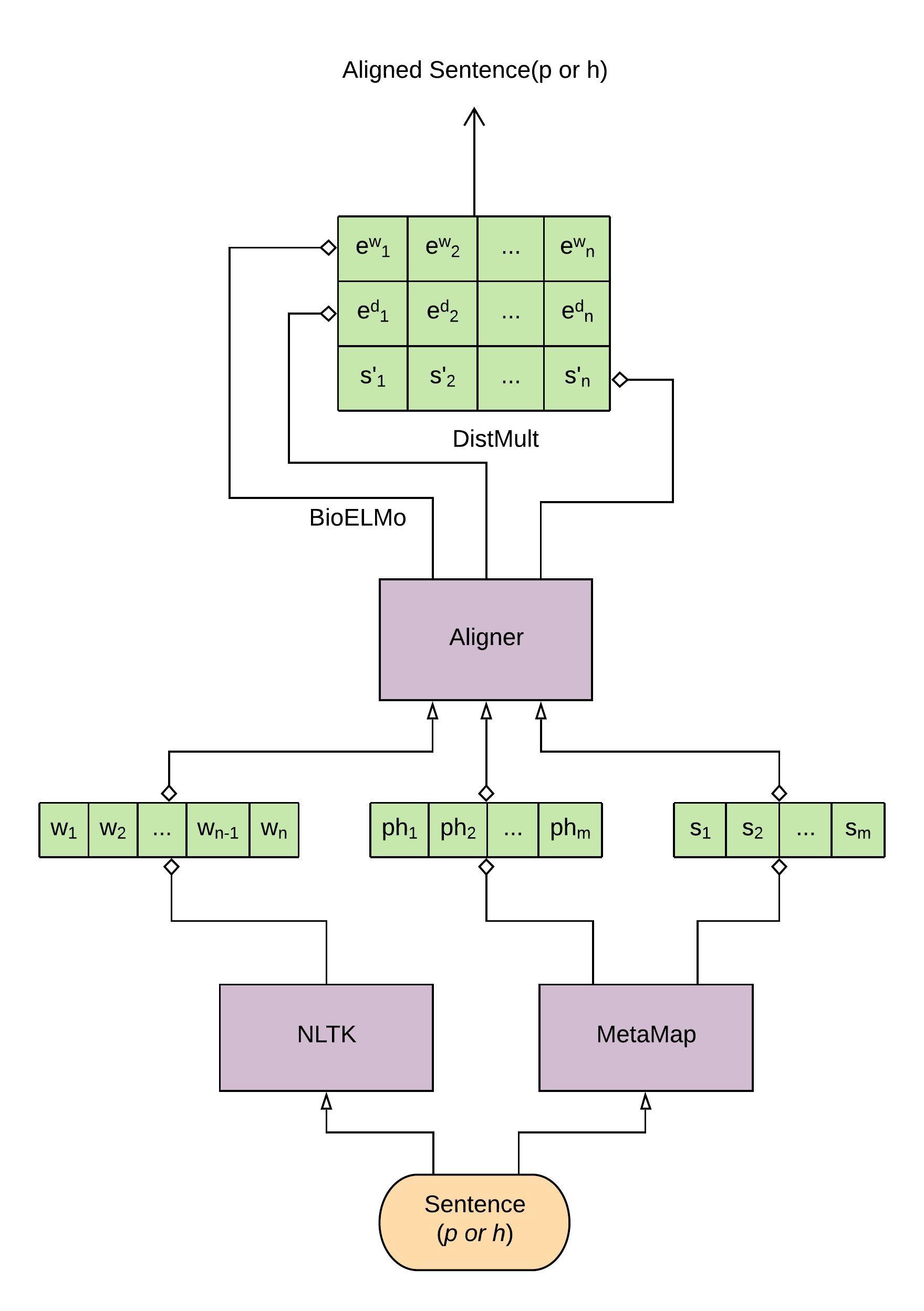}
  \vspace{-0.2cm}
  \caption{\textbf{BioELMo w/ KG pipeline} Here $w$ is the NLTK tokenized form of premise (p) or hypothesis (h), $ph$ is the MetaMap tokenized form of sentence (p or h). $s$ signifies the sentiment vector. $e^w$ and $e^d$ signify the aligned word embeddings and DistMult embeddings, respectively. $s'$ signifies the aligned sentiment vector.}
  \label{fig:model}
  \vspace{-0.5cm}
\end{figure}

\begin{figure}[t]
\centering
    \includegraphics[width=\linewidth]{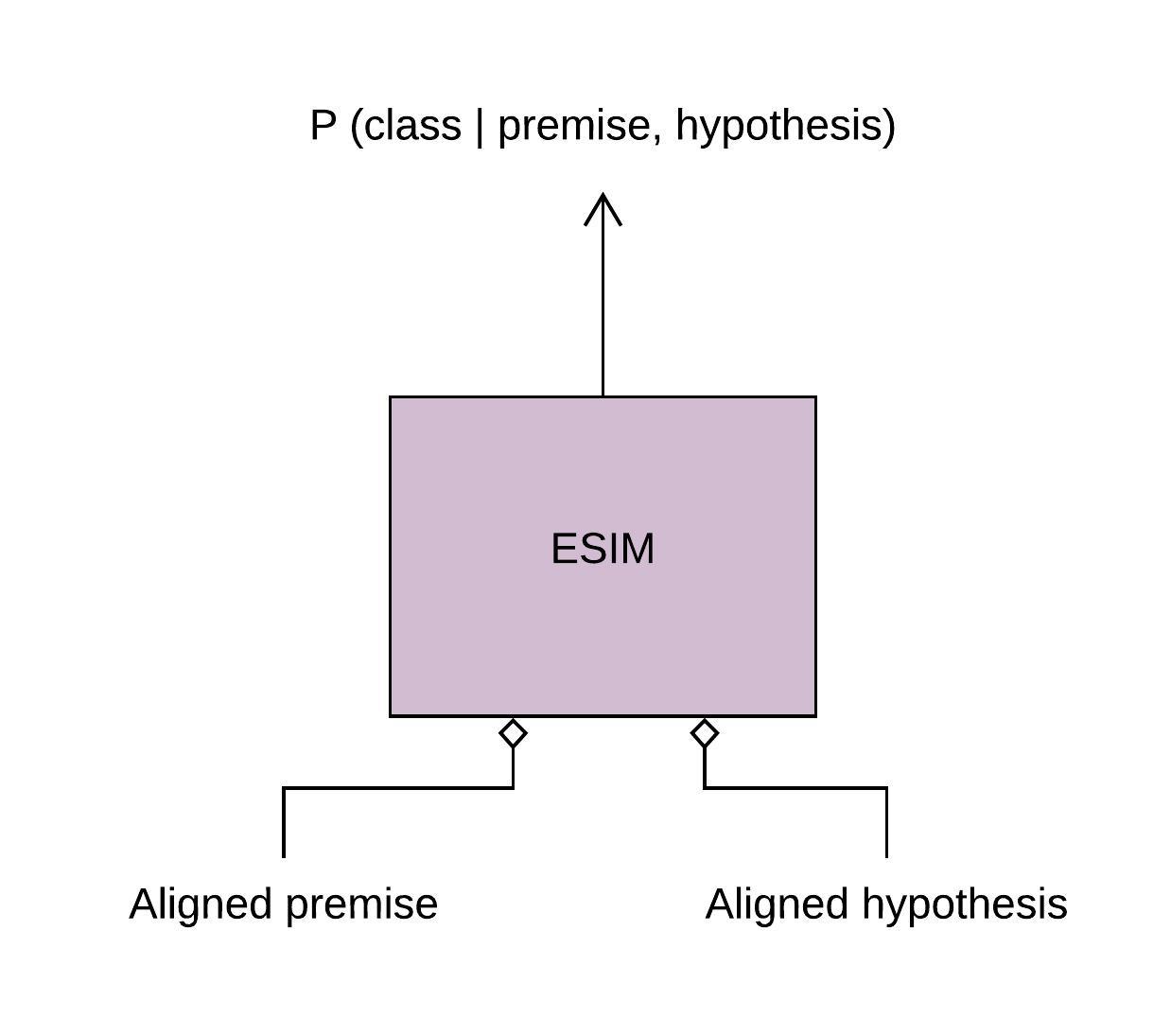}
    \vspace{-0.2cm}
    \caption{Aligned premise and hypothesis embeddings are obtained using the method described in Figure \ref{fig:model} and become inputs to the ESIM model.}
    \label{fig:esim}
  \vspace{-0.5cm}
\end{figure}

\noindent\textbf{UMLS: } Unified Medical Language System (UMLS) is a compendium which includes many health and biomedical vocabularies and standards. It provides a mapping structure between these vocabularies and is a comprehensive thesaurus and ontology of biomedical concepts. UMLS contains 3 knowledge sources: Metathesaurus, Semantic Network, and Specialist Lexicon and Lexical Tools. We use two of these sources: the Metathesaurus and the Semantic Network. The Metathesaurus comprises of over 1 million biomedical concepts and 5 million concept names. 
Each concept has numerous relationships with each other. 
Each concept in the Metathesaurus is assigned one or more Semantic Type linked to other Semantic Types through a semantic relationship. This information is provided in the Semantic Network of UMLS. There are 127 semantic types and 54 relationships in total. Semantic types include ``disease", ``symptom", ``laboratory test" and semantic relationships include ``is-a", ``part-of", ``affects". \\

\noindent\textbf{MetaMap: } MetaMap is a tool for effective entity mapping of biomedical text to the concepts and semantic type in UMLS Metathesaurus. On feeding a sentence to MetaMap, it divides the sentence into phrases based on medical concepts found in the sentence and for each medical concept it provides its unique ID in Metathesaurus, its position in the sentence, the list of semantic types the concept is mapped to, the preferred medical name and unique ID for the preferred concept (such as a concept called ``chest pain'' would be mapped to its preferred medical term ``angina''). We also get a boolean value associated with each concept denoting whether the medical concept occurs in a negative sentiment (1) or not (0). For example, in the sentence, ``The patient showed no signs of pain", the medical concept `pain' would appear with a negative sentiment. Note that, for each extracted phrase, there may be more than one related medical concepts and each concept may have more than one possible mapping. For our study, we only consider the mapping with the highest MetaMap Indexing (MMI) score, a metric provided by MetaMap. As a result, every word in a sentence has zero or one corresponding medical concept.

\textbf{Constructing the appropriate knowledge graph:} We use the MetaMap tool to process the complete MedNLI dataset and extract the relevant information from UMLS into a smaller knowledge graph.  

First, we use MetaMap to extract medical concepts from $p$ and $h$, and map them to the standard terminology in UMLS. We choose to map each medical concept to its preferred medical term. E.g., ``blood clots'' would map to ``thrombus''. This helps us to map different synonymous surface forms to the same concept. This results in 7,496 unique medical concepts from UMLS matched to various words and phrases in the MedNLI dataset. Each unique concept in UMLS becomes a node in our knowledge graph. The relations in our knowledge graph come from two sources: The Metathesaurus and the Semantic Network of UMLS. 

Using relations extracted from these two sources, we connect the filtered medical concepts from UMLS to build a smaller Knowledge Graph (subgraph of UMLS). 

We get 117,467 triples from the Metathesaurus and 23,824,105 triples from the Semantic Network, which constitute the edgelists in the prepared knowledge graph.

\textbf{Knolwedge Graph Embeddings:} To obtain the embedding from this graph, we use state-of-the-art DistMult model~\cite{distmult2015}. The choice is inspired by ~\citet{kadlec-etal-2017-knowledge}, which reports that an appropriately tuned DistMult model can produce similar or better performance while compared with the competing knowledge graph embedding models. DistMult model embeds entities (nodes) and relations (edges) as vectors. It uses a matrix dot product to measure compatibility between head ($h$) and tail ($t$) entities, connected by a relation ($r$). Logistic loss is used for training the model.

\begin{equation}
    \sigma_{DistMult}(h, r, t) = r^T(h \cdot t)
\end{equation}

\if{0}
\begin{table}[t]
\centering
\begin{tabular}{ll}
\hline
Model        & Accuracy \\ \hline
ELMo    & 75.8\%         \\
BioELMo & 78.2\% \\ 
FastText & 77.8\% \\ \hline
\end{tabular}
\caption{Baseline Results as obtained from ~\cite{jin2019probing} and ~\cite{lu2019incorporating}}
\label{table:1}
\end{table}
\fi

\textbf{Combining Knolwedge Graph Embeddings with BioELMo:}
As explained in Figure \ref{fig:model}, each sentence ($p$ or $h$) is tokenized using the simple module of NLTK\footnote{\url{https://www.nltk.org/}} as well as processed using MetaMap to get UMLS concepts. To align these, we copy the UMLS concept for a phrase to all the constituent words.\\
Once we have aligned the tokens obtained via NLTK and MetaMap, we apply BioELMo and DistMult to get the embedding vectors, \(e_{BioELMo,w}\) and \(e_{DistMult,w}\) for each word $w$. We concatenate these vectors as \(e_{w} = e_{BioELMo,w} \oplus e_{DistMult,w}\), to obtain the word representation for $w$. We call the proposed model which uses these embeddings as \textit{BioELMo w/ KG}.

\textbf{Combining Sentiment Information:} We further enhance the domain knowledge by incorporating sentiment information for a concept separately. For that purpose, we use the sentiment boolean provided by MetaMap and create a 1-d vector $(sent_w)$ containing 0 for positive medical concepts or non-medical concept and 1 for negative concept. This 1-d vector is aligned with the \(e_{DistMult,w}\) in the same fashion as explained above. We concatenate this single dimension with our concatenated resultant embeddings. Thus \(e_{w} = e_{BioELMo,w} \oplus e_{DistMult,w} \oplus sent_w\). We call the proposed model which uses these embeddings as \textit{BioELMo w/ KG + Sentiment}.

We use the vanilla ESIM model ~\cite{chen2017enhanced} and feed the obtained concatenated embeddings for each word in the premise and hypothesis to the model, to be trained for the inference task (see Figure~\ref{fig:esim}).
\begin{flushleft}
\begin{table}[!thb]
\centering
\begin{tabular}{ll}
\hline
Model        & Accuracy \\ \hline
fastText & 68.7\% \\
GloVe & 73.1\% \\
$BioELMo$~\cite{jin2019probing} & 78.2\% \\
$ESIM w/ K$~\cite{lu2019incorporating} & 77.8\% \\ 
\hline
fastText w/ KG+Sentiment & 73.67\% \\
GloVe w/ KG+Sentiment & 74.46\% \\
BioELMo w/ KG & 78.76\% \\
\textbf{BioELMo w/ KG+Sentiment}         & \textbf{79.04\%}   \\ \hline
\end{tabular}
  \vspace{-0.2cm}
\caption{Performance of our models (bottom four) along with the state-of-the-art baseline models (top four). Baseline results for fastText, GloVe are obtained from ~\citet{romanov-shivade-2018-lessons}. Adding knowledge graph information to the base models showed an absolute improvement of 4.97\% in case of fastText and 1.36\% in case of GloVe. The baseline model utilizing BioELMo as base embeddings~\cite{jin2019probing} showed an accuracy of 78.2\%. On adding knowledge graph information, we were able to improve these results to 78.76\% and on further addition of sentiment information, the accuracy rose to 79.04\%}
\label{table:2}
\end{table}
  \vspace{-0.5cm}
\end{flushleft}

\section{Experimental Results and Analysis}

As discussed earlier, we mainly consider the models presented by~\citet{jin2019probing} [$BioELMo$] and~\citet{lu2019incorporating} [\textbf{$ESIM w/ K$}] as our baselines. We report accuracy as the performance metric. Table~\ref{table:2} represents the performance comparison of our proposed models and the baselines, which shows that incorporation of knowledge graph embeddings helps to improve the model performance. All the results reported use ESIM as their base model. Further, incorporating sentiment of medical concepts gives further improvements, achieving an overall ~1\% improvement over the baseline model. 

We also see from ~\cite{jin2019probing} that BERT and BioBERT show an accuracy of 77.8\% and 81.7\%, respectively. However, they also showcase through a probing task that BioELMo is a better feature extractor than BioBERT, even though the latter has higher performance when fine tuned on MedNLI. Due to this reason, we take BioELMo as our base architecture and use our enhancements over BioELMo instead of BioBERT.

We also experimented with using fastText and GloVe as our base general embeddings. With addition of Knowledge Graph embeddings and sentiment information, the results showed an absolute improvement from 68.7\% to 73.67\% in case of fastText, and 73.1\% to 74.46\% in case of GloVe. All results are summarized in Table~\ref{table:2}.

\noindent\textbf{Training Details: }
For \textit{DistMult}, we use word embeddings dimensions to be 100. SGD was used for optimization with an initial learning rate of $10^{-4}$. The batch size was set to 100. For \textit{ESIM}, we take the dimension of hidden states of BiLSTMs to be 500. We set the dropout to 0.5 and choose an initial learning rate of $10^{-3}$. We choose a batch size of 32 and run for a maximum of 64 epochs. The training is stopped when the development loss does not decrease after 5 subsequent epochs.

\if{1}
\begin{table*}[t]
\centering
\begin{tabular}{>{\footnotesize}l>{\footnotesize}l>{\footnotesize}l>{\footnotesize}l>{\footnotesize}l}
\hline
Premise & Hypothesis & gold label & BioELMo & BioELMo w/ Knowledge & BioELMo w/ Knowledge+Sentiment\\ \hline
History of CVA. & patient has history of stroke & entailment & neutral & entailment \\
Blood glucose 626. & Patient has normal A1c & contradiction & entailment & contradiction \\
\multicolumn{1}{m{2cm}} She does admit to some diarrhea at home. & The patient is constipated & contradiction & entailment & contradiction & contradiction \\ \hline
Per report ECG with initial qtc of 410 now 475., QRS 82 initially, now 86 rate= 95. & Patient has angina & neutral & neutral & entailment \\
PAST MEDICAL HISTORY:  Type 2 diabetes mellitus. & the patient has type 1 diabetes & contradiction & entailment & entailment & entailment \\ \hline
\end{tabular}
\caption{Error Analysis}
\label{table:3}
\end{table*}

\if{0}
p: No recent change in bowel or bladder habits.
h: The patient does not have a urinary tract infection.

Example for sentiment information mattering. Here, gold label=neutral, bioelmo=entailment, bioelmo+KG=entailment, bioelmo+KG+sentiment=neutral.

Negated concepts:
p: None
h: ``Urinary Tract Infection

\fi
\subsection{Results}
Our hypothesis is that adding extra structured domain knowledge will improve the accuracy and therefore, we mainly compare BioELMo with BioELMo w/ KG (proposed model). BioELMo w/ KG achieves an accuracy of 78.76\%. All the results shown in this paper are calculated on the test dataset.

By including sentiment information obtained from MetaMap to the BioELMo w/ KG model, we achieved 79.04\% accuracy. 
\fi

\if{0}
p: The patient ruled out for a myocardial infarction; however, he was transferred to [**Hospital1 58**] for cardiac catheterization.
h: The patient did not have an acute STEMI

Example for sentiment information mattering. Here, gold label=entailment, bioelmo=entailment, bioelmo+KG=contradiction, bioelmo+KG+sentiment=entailment.

Negated concepts: 
p: ``myocardial infraction'' (heart attack)
h: ``Acute STEMI'' (very serious type of heart attack)

Only issue is that bioelmo marked it as entailment, no clue why. But can we say that since this was a direct link in KG, we can say that it was classified correctly with confidence?
\fi

\noindent\textbf{Qualitative Analysis: }We explain the efficacy of our model with the help of a few examples. Consider the sentence pair, $p$: ``History of CVA'' and $h$: ``patient has history of stroke''. In medical terms, `CVA' means `Cerebrovascular accident' which is another term for `stroke'. By Using MetaMap, we are able to find that the preferred term for `stroke' is `Cerebrovascular accident' and hence our model classified the sample pair correctly as entailment. 

In many cases, our baseline models fail to capture negative sentiment present in the premise or hypothesis. For example, in case of $p$: Watermelon stomach with gastric varices, without bleed in more than 2 years, $h$: Patient has no active bleeding, BioELMo predicts this as contradiction, whereas the gold label is entailment. But, by using the negative sentiment, captured by Metamap, for the word `bleed' in both premise and hypothesis, our model is able predict the label correctly. 


On the other hand, even though our model produces promising improvement over state-of-the-art performance, there are cases for which our model is not able to classify correctly. For the sentence pair $p$: ``She was speaking normally at that time'' and $h$: ``The patient has no known normal time where she was speaking normally.'' contradicting each other, our model predicts this to be entailment. The probable reason could be that, the negative sentiment associated with ``speaking normally'' is not captured by MetaMap and the noise in MetaMap is percolated further. 
In another example case, $p$: ``He had no EKG changes and first set of enzymes were negative.'' and $h$: ``the patient has negative enzymes.'' our model classifies this pair as entailment while the gold label is neutral. While the premise says that the first set of enzymes was negative, it gives no information about the current state. This leads us to believe that a sense of timeline is extremely important for examples like this which is not already being captured by our model. Taking care of these cases would be our immediate future work.



\section{Conclusion}
In this paper, we showed that knowledge graph embeddings obtained through applying state-of-the-art model like DistMult from UMLS could be a promising way towards incorporating domain knowledge leading to improved state-of-the-art performance for the medical NLI task. We further showed that sentiments of medical concepts can contribute to medical NLI task as well, opening a new direction to be explored further. With the emergence of knowledge graphs in different domains, the proposed approach can be tried out in other domains as well for future exploration.

\section*{Acknowledgements}
The authors would like to acknowledge the funding and support from Samsung Research Institute, Delhi (SRID), India.

\if{0}
\section{Credits}

This document has been adapted from the instructions for earlier ACL
and NAACL proceedings. It represents a recent build from
\url{https://github.com/acl-org/acl-pub}, with modifications by 
Fei Liu and Pontus Stenetorp, based on the EMNLP 2018 instructions 
by Micha Elsner and Preethi Raghavan, NAACL 2018 instructions by
Margaret Michell and Stephanie Lukin, 2017/2018 (NA)ACL bibtex
suggestions from Jason Eisner, ACL 2017 by Dan Gildea and Min-Yen Kan,
NAACL 2017 by Margaret Mitchell, ACL 2012 by Maggie Li and Michael
White, those from ACL 2010 by Jing-Shing Chang and Philipp Koehn,
1

those for ACL 2008 by Johanna D. Moore, Simone Teufel, James Allan,
and Sadaoki Furui, those for ACL 2005 by Hwee Tou Ng and Kemal
Oflazer, those for ACL 2002 by Eugene Charniak and Dekang Lin, and
earlier ACL and EACL formats.  Those versions were written by several
people, including John Chen, Henry S. Thompson and Donald
Walker. Additional elements were taken from the formatting
instructions of the {\em International Joint Conference on Artificial
  Intelligence} and the \emph{Conference on Computer Vision and
  Pattern Recognition}.

\section{Introduction}

The following instructions are directed to authors of papers submitted
to \confname{} or accepted for publication in its proceedings. All
authors are required to adhere to these specifications. Authors are
required to provide a Portable Document Format (PDF) version of their
papers. \textbf{The proceedings are designed for printing on A4
paper.}

\section{General Instructions}

Manuscripts must be in two-column format.  Exceptions to the
two-column format include the title, authors' names and complete
addresses, which must be centered at the top of the first page, and
any full-width figures or tables (see the guidelines in
Subsection~\ref{ssec:first}). {\bf Type single-spaced.}  Start all
pages directly under the top margin. See the guidelines later
regarding formatting the first page.  The manuscript should be
printed single-sided and its length
should not exceed the maximum page limit described in Section~\ref{sec:length}.
Pages are numbered for  initial submission. However, {\bf do not number 
the pages in the camera-ready version}.

By uncommenting {\small\verb|\aclfinalcopy|} at the top of this 
document, it will compile to produce an example of the camera-ready 
formatting; by leaving it commented out, the document will be anonymized 
for initial submission.

The review process is double-blind, so do not include any author information (names, addresses) when submitting a paper for review.  
However, you should maintain space for names and addresses so that they will fit in the final (accepted) version.  The \confname{} \LaTeX\ style will create a titlebox space of 2.5in for you when {\small\verb|\aclfinalcopy|} is commented out.  

The author list for submissions should include all (and only) individuals who made substantial contributions to the work presented. Each author listed on a submission to \confname{} will be notified of submissions, revisions and the final decision. No authors may be added to or removed from submissions to \confname{} after the submission deadline.

\subsection{The Ruler}
The \confname{} style defines a printed ruler which should be presented in the
version submitted for review.  The ruler is provided in order that
reviewers may comment on particular lines in the paper without
circumlocution.  If you are preparing a document without the provided
style files, please arrange for an equivalent ruler to
appear on the final output pages.  The presence or absence of the ruler
should not change the appearance of any other content on the page.  The
camera ready copy should not contain a ruler. (\LaTeX\ users may uncomment 
the {\small\verb|\aclfinalcopy|} command in the document preamble.)  

Reviewers: note that the ruler measurements do not align well with
lines in the paper -- this turns out to be very difficult to do well
when the paper contains many figures and equations, and, when done,
looks ugly. In most cases one would expect that the approximate
location will be adequate, although you can also use fractional
references ({\em e.g.}, the first paragraph on this page ends at mark $108.5$).

\subsection{Electronically-available resources}

\conforg{} provides this description in \LaTeX2e{} ({\small\tt
  emnlp-ijcnlp-2019.tex}) and PDF format ({\small\tt emnlp-ijcnlp-2019.pdf}), along
with the \LaTeX2e{} style file used to format it ({\small\tt
  emnlp-ijcnlp-2019.sty}) and an ACL bibliography style ({\small\tt
  acl\_natbib.bst}) and example bibliography ({\small\tt
  emnlp-ijcnlp-2019.bib}).  These files are all available at
{\small \url{http://emnlp-ijcnlp2019.org/downloads/emnlp-ijcnlp-2019-latex.zip}}; 
a Microsoft
Word template file ({\small\tt emnlp-ijcnlp-2019-word.docx}) and example
submission pdf ({\small\tt emnlp-ijcnlp-2019-word.pdf}) is available at
{\small \url{http://emnlp-ijcnlp2019.org/downloads/emnlp-ijcnlp-2019-word.zip}}.  We strongly
recommend the use of these style files, which have been appropriately
tailored for the \confname{} proceedings.

\subsection{Format of Electronic Manuscript}
\label{sect:pdf}

For the production of the electronic manuscript you must use Adobe's
Portable Document Format (PDF). PDF files are usually produced from
\LaTeX\ using the \textit{pdflatex} command. If your version of
\LaTeX\ produces Postscript files, you can convert these into PDF
using \textit{ps2pdf} or \textit{dvipdf}. On Windows, you can also use
Adobe Distiller to generate PDF.

Please make sure that your PDF file includes all the necessary fonts
(especially tree diagrams, symbols, and fonts with Asian
characters). When you print or create the PDF file, there is usually
an option in your printer setup to include none, all or just
non-standard fonts.  Please make sure that you select the option of
including ALL the fonts. \textbf{Before sending it, test your PDF by
  printing it from a computer different from the one where it was
  created.} Moreover, some word processors may generate very large PDF
files, where each page is rendered as an image. Such images may
reproduce poorly. In this case, try alternative ways to obtain the
PDF. One way on some systems is to install a driver for a postscript
printer, send your document to the printer specifying ``Output to a
file'', then convert the file to PDF.

It is of utmost importance to specify the \textbf{A4 format} (21 cm
x 29.7 cm) when formatting the paper. When working with
{\tt dvips}, for instance, one should specify {\tt -t a4}.
Or using the command \verb|\special{papersize=210mm,297mm}| in the latex
preamble (directly below the \verb|\usepackage| commands). Then using 
{\tt dvipdf} and/or {\tt pdflatex} which would make it easier for some.

Print-outs of the PDF file on A4 paper should be identical to the
hardcopy version. If you cannot meet the above requirements about the
production of your electronic submission, please contact the
publication chairs as soon as possible.

\subsection{Layout}
\label{ssec:layout}

Format manuscripts two columns to a page, in the manner these
instructions are formatted. The exact dimensions for a page on A4
paper are:

\begin{itemize}
\item Left and right margins: 2.5 cm
\item Top margin: 2.5 cm
\item Bottom margin: 2.5 cm
\item Column width: 7.7 cm
\item Column height: 24.7 cm
\item Gap between columns: 0.6 cm
\end{itemize}

\noindent Papers should not be submitted on any other paper size.
 If you cannot meet the above requirements about the production of 
 your electronic submission, please contact the publication chairs 
 above as soon as possible.

\subsection{Fonts}

For reasons of uniformity, Adobe's {\bf Times Roman} font should be
used. In \LaTeX2e{} this is accomplished by putting

\begin{quote}
\begin{verbatim}
\usepackage{times}
\usepackage{latexsym}
\end{verbatim}
\end{quote}
in the preamble. If Times Roman is unavailable, use {\bf Computer
  Modern Roman} (\LaTeX2e{}'s default).  Note that the latter is about
  10\% less dense than Adobe's Times Roman font.

\begin{table}[t!]
\begin{center}
\begin{tabular}{|l|rl|}
\hline \bf Type of Text & \bf Font Size & \bf Style \\ \hline
paper title & 15 pt & bold \\
author names & 12 pt & bold \\
author affiliation & 12 pt & \\
the word ``Abstract'' & 12 pt & bold \\
section titles & 12 pt & bold \\
document text & 11 pt  &\\
captions & 10 pt & \\
abstract text & 10 pt & \\
bibliography & 10 pt & \\
footnotes & 9 pt & \\
\hline
\end{tabular}
\end{center}
\caption{\label{font-table} Font guide. }
\end{table}

\subsection{The First Page}
\label{ssec:first}

Center the title, author's name(s) and affiliation(s) across both
columns. Do not use footnotes for affiliations. Do not include the
paper ID number assigned during the submission process. Use the
two-column format only when you begin the abstract.

{\bf Title}: Place the title centered at the top of the first page, in
a 15-point bold font. (For a complete guide to font sizes and styles,
see Table~\ref{font-table}) Long titles should be typed on two lines
without a blank line intervening. Approximately, put the title at 2.5
cm from the top of the page, followed by a blank line, then the
author's names(s), and the affiliation on the following line. Do not
use only initials for given names (middle initials are allowed). Do
not format surnames in all capitals ({\em e.g.}, use ``Mitchell'' not
``MITCHELL'').  Do not format title and section headings in all
capitals as well except for proper names (such as ``BLEU'') that are
conventionally in all capitals.  The affiliation should contain the
author's complete address, and if possible, an electronic mail
address. Start the body of the first page 7.5 cm from the top of the
page.

The title, author names and addresses should be completely identical
to those entered to the electronical paper submission website in order
to maintain the consistency of author information among all
publications of the conference. If they are different, the publication
chairs may resolve the difference without consulting with you; so it
is in your own interest to double-check that the information is
consistent.

{\bf Abstract}: Type the abstract at the beginning of the first
column. The width of the abstract text should be smaller than the
width of the columns for the text in the body of the paper by about
0.6 cm on each side. Center the word {\bf Abstract} in a 12 point bold
font above the body of the abstract. The abstract should be a concise
summary of the general thesis and conclusions of the paper. It should
be no longer than 200 words. The abstract text should be in 10 point font.

{\bf Text}: Begin typing the main body of the text immediately after
the abstract, observing the two-column format as shown in 
the present document. Do not include page numbers.

{\bf Indent}: Indent when starting a new paragraph, about 0.4 cm. Use 11 points for text and subsection headings, 12 points for section headings and 15 points for the title.

\begin{table}
\centering
\small
\begin{tabular}{cc}
\begin{tabular}{|l|l|}
\hline
{\bf Command} & {\bf Output}\\\hline
\verb|{\"a}| & {\"a} \\
\verb|{\^e}| & {\^e} \\
\verb|{\`i}| & {\`i} \\ 
\verb|{\.I}| & {\.I} \\ 
\verb|{\o}| & {\o} \\
\verb|{\'u}| & {\'u}  \\ 
\verb|{\aa}| & {\aa}  \\\hline
\end{tabular} & 
\begin{tabular}{|l|l|}
\hline
{\bf Command} & {\bf  Output}\\\hline
\verb|{\c c}| & {\c c} \\ 
\verb|{\u g}| & {\u g} \\ 
\verb|{\l}| & {\l} \\ 
\verb|{\~n}| & {\~n} \\ 
\verb|{\H o}| & {\H o} \\ 
\verb|{\v r}| & {\v r} \\ 
\verb|{\ss}| & {\ss} \\\hline
\end{tabular}
\end{tabular}
\caption{Example commands for accented characters, to be used in, {\em e.g.}, \BibTeX\ names.}\label{tab:accents}
\end{table}

\subsection{Sections}

{\bf Headings}: Type and label section and subsection headings in the
style shown on the present document.  Use numbered sections (Arabic
numerals) in order to facilitate cross references. Number subsections
with the section number and the subsection number separated by a dot,
in Arabic numerals.
Do not number subsubsections.

\begin{table*}[t!]
\centering
\begin{tabular}{lll}
  output & natbib & previous \conforg{} style files\\
  \hline
  \citep{Gusfield:97} & \verb|\citep| & \verb|\cite| \\
  \citet{Gusfield:97} & \verb|\citet| & \verb|\newcite| \\
  \citeyearpar{Gusfield:97} & \verb|\citeyearpar| & \verb|\shortcite| \\
\end{tabular}
\caption{Citation commands supported by the style file.
  The citation style is based on the natbib package and
  supports all natbib citation commands.
  It also supports commands defined in previous \conforg{} style files
  for compatibility.
  }
\end{table*}

{\bf Citations}: Citations within the text appear in parentheses
as~\cite{Gusfield:97} or, if the author's name appears in the text
itself, as Gusfield~\shortcite{Gusfield:97}.
Using the provided \LaTeX\ style, the former is accomplished using
{\small\verb|\cite|} and the latter with {\small\verb|\shortcite|} or {\small\verb|\newcite|}. Collapse multiple citations as in~\cite{Gusfield:97,Aho:72}; this is accomplished with the provided style using commas within the {\small\verb|\cite|} command, {\em e.g.}, {\small\verb|\cite{Gusfield:97,Aho:72}|}. Append lowercase letters to the year in cases of ambiguities.  
 Treat double authors as
in~\cite{Aho:72}, but write as in~\cite{Chandra:81} when more than two
authors are involved. Collapse multiple citations as
in~\cite{Gusfield:97,Aho:72}. Also refrain from using full citations
as sentence constituents.

We suggest that instead of
\begin{quote}
  ``\cite{Gusfield:97} showed that ...''
\end{quote}
you use
\begin{quote}
``Gusfield \shortcite{Gusfield:97}   showed that ...''
\end{quote}

If you are using the provided \LaTeX{} and Bib\TeX{} style files, you
can use the command \verb|\citet| (cite in text)
to get ``author (year)'' citations.

If the Bib\TeX{} file contains DOI fields, the paper
title in the references section will appear as a hyperlink
to the DOI, using the hyperref \LaTeX{} package.
To disable the hyperref package, load the style file
with the \verb|nohyperref| option: \\
{\small \verb|\usepackage[nohyperref]{emnlp-ijcnlp-2019}|}\\

\textbf{Digital Object Identifiers}: As part of our work to make ACL
materials more widely used and cited outside of our discipline, ACL
has registered as a CrossRef member, as a registrant of Digital Object
Identifiers (DOIs), the standard for registering permanent URNs for
referencing scholarly materials. 
As of 2017, we are requiring all camera-ready references to contain the appropriate DOIs (or as a second resort, the hyperlinked ACL Anthology Identifier) to all cited works. Thus, please ensure that you use References that contain DOI or URLs for any of the ACL materials that you reference. 
Appropriate records should be found
for most materials in the current ACL Anthology at
\url{http://aclanthology.info/}.

As examples, we cite \cite{P16-1001} to show you how papers with a DOI
will appear in the bibliography.  We cite \cite{C14-1001} to show how
papers without a DOI but with an ACL Anthology Identifier will appear
in the bibliography.  

\textbf{Anonymity:} As reviewing will be double-blind, the submitted
version of the papers should not include the authors' names and
affiliations. Furthermore, self-references that reveal the author's
identity, {\em e.g.},
\begin{quote}
``We previously showed \cite{Gusfield:97} ...''  
\end{quote}
should be avoided. Instead, use citations such as 
\begin{quote}
``\citeauthor{Gusfield:97} \shortcite{Gusfield:97}
previously showed ... ''
\end{quote}

Preprint servers such as arXiv.org and workshops that do not
have published proceedings are not considered archival for purposes of
submission. However, to preserve the spirit of blind review, authors
are encouraged to refrain from posting until the completion of the
review process. Otherwise, authors must state in the online submission
form the name of the workshop or preprint server and title of the
non-archival version. The submitted version should be suitably
anonymized and not contain references to the prior non-archival
version. Reviewers will be told: ``The author(s) have notified us that
there exists a non-archival previous version of this paper with
significantly overlapping text. We have approved submission under
these circumstances, but to preserve the spirit of blind review, the
current submission does not reference the non-archival version.''

\textbf{Please do not use anonymous citations} and do not include
 when submitting your papers. Papers that do not
conform to these requirements may be rejected without review.

\textbf{References}: Gather the full set of references together under
the heading {\bf References}; place the section before any Appendices,
unless they contain references. Arrange the references alphabetically
by first author, rather than by order of occurrence in the text.
By using a .bib file, as in this template, this will be automatically 
handled for you. See the \verb|\bibliography| commands near the end for more.

Provide as complete a citation as possible, using a consistent format,
such as the one for {\em Computational Linguistics\/} or the one in the 
{\em Publication Manual of the American 
Psychological Association\/}~\cite{APA:83}. Use of full names for
authors rather than initials is preferred. A list of abbreviations
for common computer science journals can be found in the ACM 
{\em Computing Reviews\/}~\cite{ACM:83}.

The \LaTeX{} and Bib\TeX{} style files provided roughly fit the
American Psychological Association format, allowing regular citations, 
short citations and multiple citations as described above.  

\begin{itemize}
\item Example citing an arxiv paper: \cite{rasooli-tetrault-2015}. 
\item Example article in journal citation: \cite{Ando2005}.
\item Example article in proceedings, with location: \cite{borsch2011}.
\item Example article in proceedings, without location: \cite{andrew2007scalable}.
\end{itemize}
See corresponding .bib file for further details.

Submissions should accurately reference prior and related work, including code and data. If a piece of prior work appeared in multiple venues, the version that appeared in a refereed, archival venue should be referenced. If multiple versions of a piece of prior work exist, the one used by the authors should be referenced. Authors should not rely on automated citation indices to provide accurate references for prior and related work.

{\bf Appendices}: Appendices, if any, directly follow the text and the
references (but see above).  Letter them in sequence and provide an
informative title: {\bf Appendix A. Title of Appendix}.

\subsection{URLs}

URLs can be typeset using the \verb|\url| command. However, very long
URLs cause a known issue in which the URL highlighting may incorrectly
cross pages or columns in the document. Please check carefully for
URLs too long to appear in the column, which we recommend you break,
shorten or place in footnotes. Be aware that actual URL should appear
in the text in human-readable format; neither internal nor external
hyperlinks will appear in the proceedings.

\subsection{Footnotes}

{\bf Footnotes}: Put footnotes at the bottom of the page and use 9
point font. They may be numbered or referred to by asterisks or other
symbols.\footnote{This is how a footnote should appear.} Footnotes
should be separated from the text by a line.\footnote{Note the line
separating the footnotes from the text.}

\subsection{Graphics}

{\bf Illustrations}: Place figures, tables, and photographs in the
paper near where they are first discussed, rather than at the end, if
possible.  Wide illustrations may run across both columns.  Color
illustrations are discouraged, unless you have verified that  
they will be understandable when printed in black ink.

{\bf Captions}: Provide a caption for every illustration; number each one
sequentially in the form:  ``Figure 1. Caption of the Figure.'' ``Table 1.
Caption of the Table.''  Type the captions of the figures and 
tables below the body, using 11 point text.

\subsection{Accessibility}
\label{ssec:accessibility}

In an effort to accommodate people who are color-blind (as well as those printing
to paper), grayscale readability for all accepted papers will be
encouraged.  Color is not forbidden, but authors should ensure that
tables and figures do not rely solely on color to convey critical
distinctions. A simple criterion: All curves and points in your figures should be clearly distinguishable without color.




\section{Translation of non-English Terms}

It is also advised to supplement non-English characters and terms
with appropriate transliterations and/or translations
since not all readers understand all such characters and terms.
Inline transliteration or translation can be represented in
the order of: original-form transliteration ``translation''.

\section{Length of Submission}
\label{sec:length}

The \confname{} main conference accepts submissions of long papers and
short papers.
 Long papers may consist of up to eight (8) pages of
content plus unlimited pages for references. Upon acceptance, final
versions of long papers will be given one additional page -- up to nine (9)
pages of content plus unlimited pages for references -- so that reviewers' comments
can be taken into account. Short papers may consist of up to four (4)
pages of content, plus unlimited pages for references. Upon
acceptance, short papers will be given five (5) pages in the
proceedings and unlimited pages for references. 

For both long and short papers, all illustrations and tables that are part
of the main text must be accommodated within these page limits, observing
the formatting instructions given in the present document. Supplementary
material in the form of appendices does not count towards the page limit; see appendix A for further information.

However, note that supplementary material should be supplementary
(rather than central) to the paper, and that reviewers may ignore
supplementary material when reviewing the paper (see Appendix
\ref{sec:supplemental}). Papers that do not conform to the specified
length and formatting requirements are subject to be rejected without
review.

Workshop chairs may have different rules for allowed length and
whether supplemental material is welcome. As always, the respective
call for papers is the authoritative source.

\fi
\if{0}

\section*{Acknowledgments}

The acknowledgments should go immediately before the references.  Do \cite{andrew2007scalable}.
not number the acknowledgments section. Do not include this section
when submitting your paper for review. \\

\noindent {\bf Preparing References:} \\

Include your own bib file like this:
{\small\verb|\bibliographystyle{acl_natbib}|
\verb|\bibliography{emnlp-ijcnlp-2019}|}

Where \verb|emnlp-ijcnlp-2019| corresponds to the {\tt emnlp-ijcnlp-2019.bib} file.

\fi

\bibliography{emnlp-ijcnlp-2019}
\bibliographystyle{acl_natbib}

\if{0}
\appendix
\input{emnlp-ijcnlp-2019-latex/Appendix.tex}
\fi
\end{document}